\documentclass[runningheads]{llncs}

\usepackage{eccv}

\usepackage{eccvabbrv}

\usepackage{graphicx}
\usepackage{booktabs}
\usepackage[utf8]{inputenc}
\usepackage{newunicodechar}
\newunicodechar{，}{,}
\usepackage{amsmath,amssymb}
\usepackage{multirow}
\usepackage{algorithm}
\usepackage{algorithmic}
\usepackage{xcolor}
\usepackage{subcaption}
\usepackage{bm}
\usepackage{colortbl}
\usepackage{subcaption}
\usepackage{graphicx}

\newcommand{\method}{IGAR}

\newcommand{\tmark}[1]{\cellcolor{red!15}{#1}}
\newcommand{\gmark}[1]{\cellcolor{green!12}{#1}}

\usepackage{hyperref}

\usepackage{orcidlink}

\begin{document}

\title{Restoring Linguistic Grounding in VLA Models via Train-Free Attention Recalibration}

\titlerunning{IGAR}

\author{Ninghao Zhang \inst{1}\textsuperscript{*} \orcidlink{0009-0005-1182-4587} \and
Bin Zhu \inst{2}\textsuperscript{*}\textsuperscript{\textdagger} \orcidlink{0000-0002-9213-2611} \and
Shijie Zhou \inst{3, 4} \and
Jingjing Chen \inst{3, 4}\orcidlink{0000-0003-3148-264X}}
\authorrunning{N.~Zhang et al.}

\institute{\textsuperscript{1} Tsinghua University
\textsuperscript{2} Singapore Management University \\
\textsuperscript{3} Institute of Trustworthy Embodied AI, Fudan University\\
\textsuperscript{4} Shanghai Key Laboratory of Multimodal Embodied AI \\
Project page: \url{https://ray-nh.github.io/igar/}
}

\maketitle

\footnotetext[1]{\textsuperscript{*} Equal contribution.}
\footnotetext[2]{\textsuperscript{\textdagger} Corresponding author and project lead. Email: binzhu@smu.edu.sg}

\begin{abstract}
Vision-Language-Action (VLA) models enable robots to perform manipulation tasks directly from natural language instructions and are increasingly viewed as a foundation for generalist robotic policies. However, their reliability under Out-Of-Distribution (OOD) instructions remains underexplored. In this paper, we reveal a critical failure mode in which VLA policies continue executing visually plausible actions even when the language instruction contradicts the scene. We refer to this phenomenon as \textbf{linguistic blindness}, where VLA policies prioritize visual priors over instruction semantics during action generation. To systematically analyze this issue, we introduce \textbf{ICBench}, a diagnostic benchmark constructed from the LIBERO dataset that probes language–action coupling by injecting controlled OOD instruction contradictions while keeping the visual environment unchanged. Evaluations on three representative VLA architectures, including $\pi_0$, $\pi_{0.5}$, and OpenVLA-OFT, show that these models frequently succeed at tasks despite logically impossible instructions, revealing a strong visual bias in action generation. 
To mitigate this issue, we propose \textbf{Instruction-Guided Attention Recalibration (IGAR)}, a train-free inference-time mechanism that rebalances attention distributions to restore the influence of language instructions. IGAR operates without retraining or architectural modification and can be directly applied to existing VLA models. Experiments across 30 LIBERO tasks demonstrate that IGAR substantially reduces erroneous execution under OOD contradictory instructions while preserving baseline task performance. We additionally validate the approach on a real Franka robotic arm, where IGAR effectively prevents manipulation triggered by inconsistent instructions.

  \keywords{Vision-Language-Action Models \and Robotic Manipulation \and Attention Recalibration}
\end{abstract}

\section{Introduction}
\label{sec:intro}

\begin{figure}[htb]
    \centering
    \includegraphics[width=0.95\textwidth]{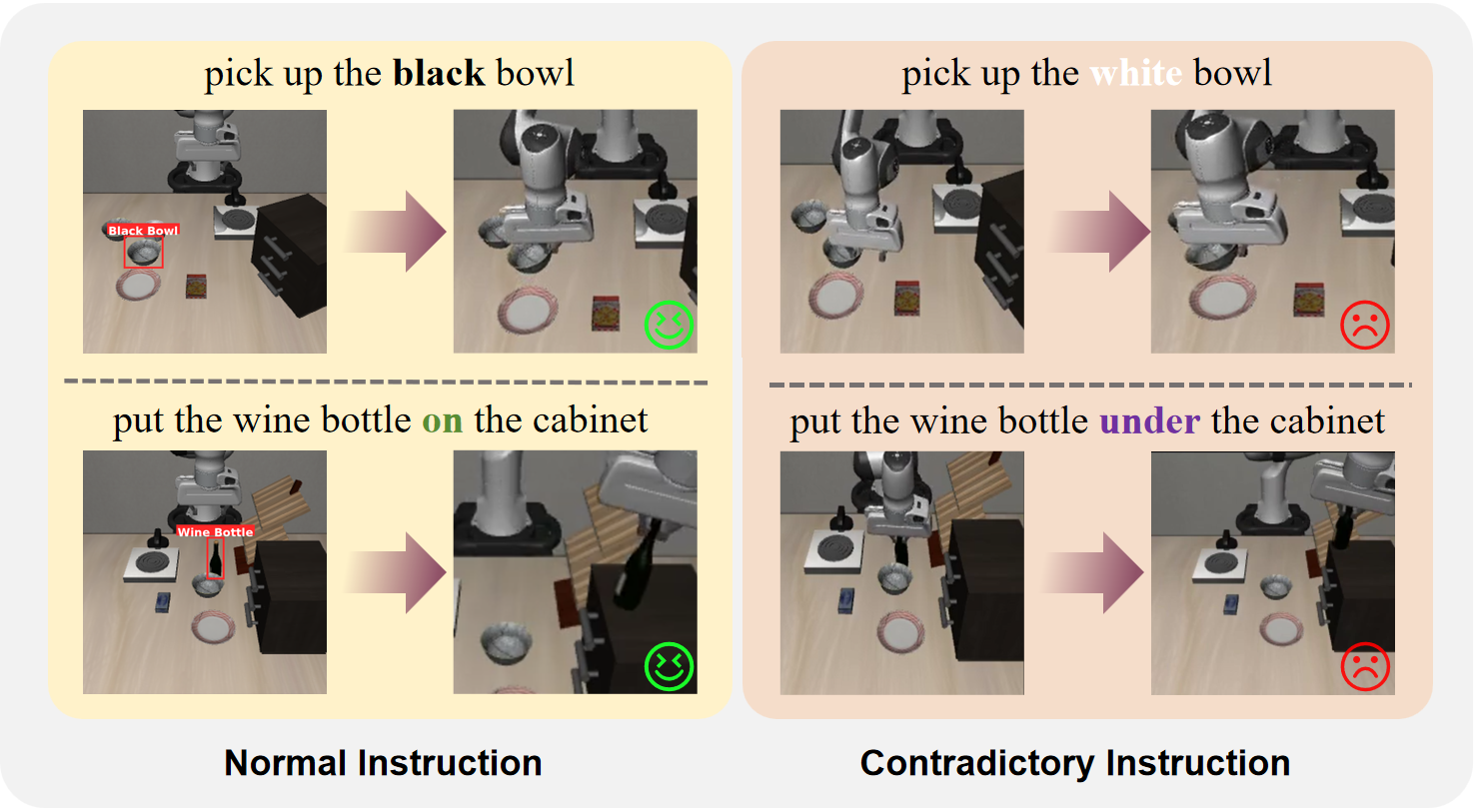}
    \caption{\textbf{Linguistic blindness in Vision-Language-Action (VLA) models.} Under normal instructions (left), the robot completes the task correctly. Under contradictory instructions (right), a structured form of OOD linguistic input, the robot often follows the same visually plausible trajectory while ignoring the instruction.}
    \label{fig:motivation}
\end{figure}

Vision-Language-Action (VLA) models are emerging as a promising paradigm for building generalist robotic policies. By integrating large-scale vision-language models~\cite{team2023gemini, beyer2024paligemma, touvron2023llama, driess2023palm} with action generation modules, these systems enable robots to execute complex manipulation tasks directly from natural language instructions across diverse environments~\cite{rt2, pi0, pi05, openvla, openvla_oft, team2025gemini, open_x_embodiment, vlabench}.

Despite these advances, the reliability of VLA models in real-world deployment remains a critical concern. In safety-critical environments, robots must strictly obey linguistic constraints provided by users. However, we observe that modern VLA models frequently execute visually plausible actions even when the instruction is semantically inconsistent with the scene. As illustrated in Fig.~\ref{fig:motivation}, a robot can successfully execute tasks such as \emph{``pick up the black bowl''} or \emph{``put the wine bottle on the cabinet''} once a VLA model is well trained. However, when the instruction is modified to contradict the scene, for example, \emph{``pick up the white bowl''} when no white bowl is present, or \emph{``put the wine bottle under the cabinet''} when this is physically impossible, the robot often continues to execute the original visually plausible trajectory while ignoring the instruction. These contradictory instructions represent a structured form of out-of-distribution (OOD) linguistic input, revealing a fundamental failure mode that we term \textbf{linguistic blindness}: the tendency of VLA policies to prioritize visual priors over instruction semantics during action generation. This vulnerability poses serious risks for real-world robotics. Unlike conversational AI systems, errors in robotic control directly translate into physical actions that may damage objects, violate safety constraints, or cause hazardous behavior~\cite{guiochet2017safety, zacharaki2020safety, team2025evaluating, li2026robotrustbench}. Ensuring that robots remain sensitive to language instructions is therefore essential for trustworthy embodied intelligence~\cite{zhang2025safevla, zhou2026rc}.

However, diagnosing linguistic grounding failures in VLA models is challenging. Existing evaluations primarily measure task success under valid instructions~\cite{libero, khazatsky2024droid, li2023behavior}, which cannot distinguish whether successful execution arises from genuine language grounding or from purely visual heuristics. To address this limitation, we introduce \textbf{ICBench}, a controlled diagnostic benchmark built upon the LIBERO dataset~\cite{libero}. ICBench isolates the coupling between language and action by injecting structured instruction contradictions while keeping the visual scene unchanged. Specifically, we modify task instructions by altering object attributes or spatial relations of the target location, as shown in Fig.~\ref{fig:motivation}. These modifications create semantically inconsistent instructions that function as controlled OOD linguistic perturbations. Under such contradictions, a language-grounded policy should detect the inconsistency and fail to execute the task, whereas a linguistically insensitive policy may still succeed by relying on visual priors.
Using ICBench, we conduct a systematic analysis of three representative VLA architectures: $\pi_0$~\cite{pi0}, $\pi_{0.5}$~\cite{pi05}, and OpenVLA-OFT~\cite{openvla_oft}. Our evaluation reveals a striking phenomenon: across multiple task suites, VLA models often maintain high success rates even when instructions are logically impossible or inconsistent. This demonstrates that action generation is frequently dominated by visual cues, with language playing only a limited role in guiding behavior.

To address this issue, we propose \textbf{I}nstruction-\textbf{G}uided \textbf{A}ttention \textbf{R}ecalibration (\textbf{IGAR}), a train-free intervention that restores linguistic grounding by rebalancing attention distributions during the forward pass. IGAR identifies attention sink~\cite{xiaoefficient, kangsee} tokens through hidden-state spike analysis, selects grounding-critical cross-modal heads, and redistributes attention mass toward under-weighted instruction tokens. Importantly, IGAR requires no gradient updates, no additional training data, and no architectural modification, making it a lightweight plug-and-play module for deployed robotic VLA policies. Extensive experiments on 30 tasks from the LIBERO benchmark demonstrate the effectiveness of IGAR. Under OOD contradictory instructions, IGAR significantly reduces erroneous task execution and substantially increases the Linguistic Grounding Score (LGS), indicating stronger reliance on instruction semantics. At the same time, IGAR preserves VLA baseline task performance under normal instructions. We further validate our approach on a real Franka robotic arm, where IGAR successfully interrupts the manipulation tasks when contradictory instructions are applied.

\section{Related Work}
\label{sec:related}

\subsection{Vision-Language-Action Models}
The integration of Vision Language Models (VLMs) with robotic control has accelerated the transition from task-specific policies to generalist Vision-Language-Action (VLA) models. Early milestones like RT-1~\cite{rt1} and RT-2~\cite{rt2} formulated motor control as sequence-to-sequence generation, a paradigm rapidly scaled by massive robotic datasets~\cite{open_x_embodiment}. Subsequent models further improve action generation through stronger vision-language backbones and advanced control paradigms. For example, OpenVLA~\cite{openvla} leverages large pretrained VLMs, while architectures like Octo~\cite{octo}, $\pi_0$~\cite{pi0}, OpenVLA-OFT~\cite{openvla_oft}, RDT~\cite{RDT}, GR00T~\cite{bjorck2025gr00t}, and $\pi_{0.5}$~\cite{pi05} refine continuous control through diffusion and flow-matching paradigms. Despite these advances, the internal mechanisms fusing visual and textual modalities during action generation remain poorly understood, particularly regarding how language instructions influence control decisions.

\subsection{Linguistic Grounding and Modality Bias}
Multimodal foundation models are known to exhibit strong modality biases that can undermine reliable language grounding. 
Multimodal foundation models are vulnerable to modality collapse and visual hallucination. Evaluations~\cite{pope} show Vision-Language Models (VLMs) frequently hallucinate absent objects due to strong textual priors. Conversely, VLMs often exhibit bag-of-words behavior or disregard order and composition information in text~\cite{vl_bias}. Recent works on Gaslighting Negation Attack~\cite{zhu2026benchmarking, zhu2026benchmarking-reason, tang2026spatiotemporal} further demonstrate that negation-based instructions can systematically mislead multimodal large models, revealing their limited ability to faithfully ground contradictory language. Existing works have also identified attention imbalance in LLMs and VLMs, where certain visual tokens attract disproportionately high attention regardless of their relevance to the instruction~\cite{kangsee, xiaoefficient}. This vulnerability becomes critical in VLA architectures: unlike conversational agents, ignoring linguistically grounded constraints in robotic control can lead to catastrophic physical failures~\cite{guiochet2017safety}.

Recognizing this risk, initial efforts have sought to address linguistic vulnerabilities in embodied control. To mitigate text-following biases, data-centric strategies like CAST~\cite{cast} and CounterfactualVLA~\cite{counterfactualvla} expand training corpora with synthesized counterfactual scenarios, particularly within navigation and autonomous driving. SayCan~\cite{brohan2023can} grounds language in robotic affordances by combining large language models with value functions of low-level skills for real-world action selection. In addition, the community has also started probing policy resilience, such as LIBERO-PLUS~\cite{liberoplus}, which evaluates robustness against perturbations such as camera viewpoints and instructions. However, most existing works focus on task success under valid instructions and do not explicitly disentangle whether success arises from genuine language grounding or from visually driven heuristics. In contrast, this paper introduces a controlled diagnostic setting that isolates language–action coupling by injecting structured contradictory instructions while keeping the visual scene unchanged.  Unlike existing attention-sink mitigation methods~\cite{xiaoefficient, kangsee, jiao2025don} developed for language generation in LLMs or MLLMs, IGAR is the first inference-time intervention designed for VLA policies, selectively recalibrating grounding-critical cross-modal attention to recover instruction influence during robot action generation.

\section{ICBench: A Controlled Instruction Contradiction Benchmark}
\label{Sec:ICBench}
To rigorously evaluate linguistic grounding in Vision-Language-Action (VLA) models, we introduce ICBench, a controlled benchmark that injects out-of-distribution (OOD) semantic contradictions into task instructions while keeping the visual scene and environment dynamics unchanged. The objective of ICBench is diagnostic rather than adversarial: it probes whether language meaningfully influences action generation. If a model genuinely integrates linguistic constraints into motor planning, contradictory instructions should prevent successful execution. Conversely, if the policy predominantly relies on visual priors, it may continue executing visually valid behaviors despite semantic inconsistencies. 

\subsection{Instruction Contradiction Construction}
Let a standard task be defined by instruction $\ell$ and environment observation $\bm{o}_t$. ICBench constructs a modified instruction $\tilde{\ell}$ such that: (1) The environment observation $\bm{o}_t$ remains unchanged. (2) $\tilde{\ell}$ is semantically incompatible with the scene configuration. By holding the environment fixed, ICBench enables controlled measurement of language–action coupling without confounding perception or control quality.

Under this setting, the contradictory instruction $\tilde{\ell}$ can be viewed as an OOD linguistic perturbation relative to the original task instruction. A language-grounded policy should fail or abstain from execution, while a linguistically insensitive policy may still succeed by exploiting visual priors. Importantly, within ICBench, high task success under contradictory instructions indicates weak linguistic grounding. Therefore, task success rate (SR) serves as a proxy for instruction adherence rather than manipulation capability.

\subsection{Contradiction Taxonomy and Design Principles}
\noindent \textbf{Contradiction Taxonomy.} ICBench introduces four structured contradiction types. Each perturbation modifies only a minimal subset of instruction tokens while preserving grammatical plausibility and fluency. These modifications introduce controlled OOD linguistic inputs without altering the visual environment.

\textit{V1: Operand Attribute Substitution.} The manipulated object’s descriptive attribute (e.g., colour) is replaced with a non-existent alternative. For instance, “pick up the black bowl” → “pick up the white bowl”. This tests grounding of object-level attribute semantics.

\textit{V2: Target Attribute Augmentation.} A contradictory attribute is inserted into the target location description. For example, “place it on the plate” → “place it on the black plate.” This evaluates grounding of relational destination constraints.

\textit{V3: Dual Attribute Perturbation.} Both operand and target attributes are contradicted simultaneously. For example, “put black bowl on the plate” → “put white bowl on black plate”. This increases semantic inconsistency while maintaining syntactic validity.

\textit{V4: Spatial Relation Substitution.} The spatial preposition is replaced with a contradictory alternative. For instance, “put the block on the table” → “put the block under the table”. Unlike attribute substitutions, spatial contradictions directly alter trajectory planning, thereby testing deeper relational grounding.

\noindent \textbf{Design Principles.} ICBench is designed with three key principles: (1) \textit{Minimal Surface Perturbation.} Only a small number of tokens are modified to ensure the instruction remains natural and close to the training distribution. (2) \textit{Deterministic Unsatisfiability.} Each contradictory instruction is guaranteed to be incompatible with the fixed scene configuration, eliminating ambiguous evaluation cases. (3) \textit{Different Semantic Complexity.} Perturbations range from local attribute mismatches (V1–V3) to spatial inconsistencies (V4), enabling analysis of shallow versus deep linguistic grounding.

\begin{figure}[H]
    \centering
    \includegraphics[width=\textwidth]{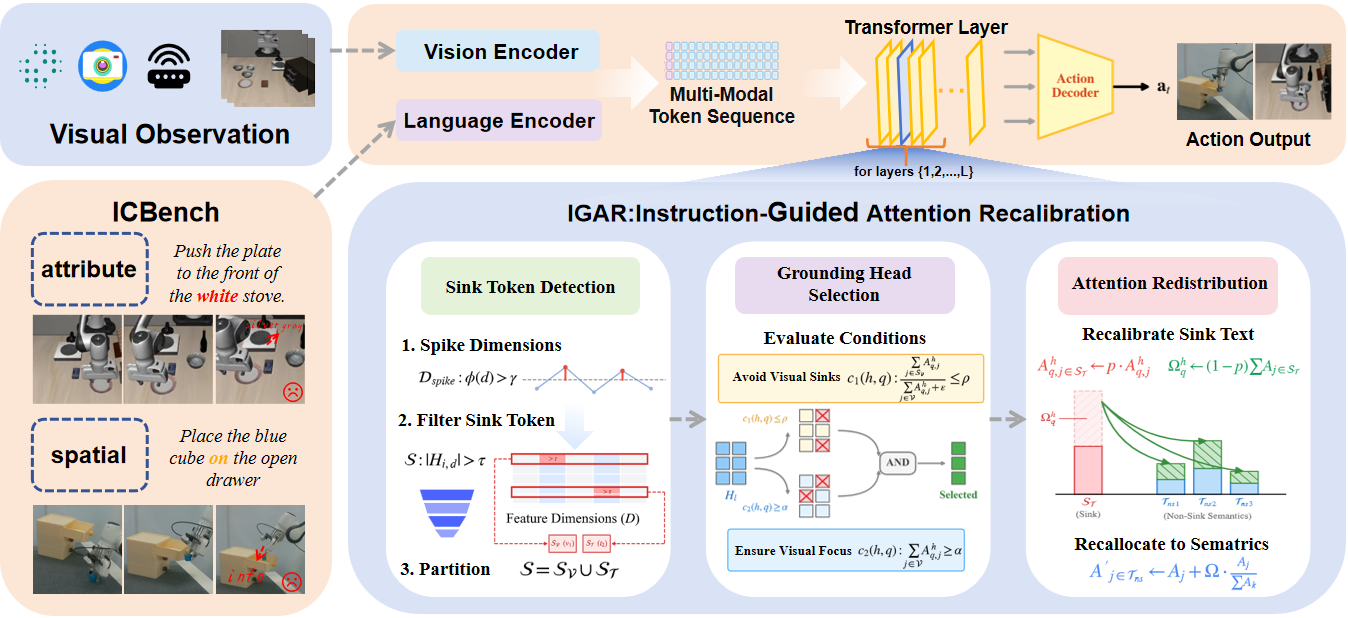}
    \caption{\textbf{Overview of the IGAR framework.} IGAR is a train-free and plug-and-play intervention that restores linguistic grounding in VLA models via three stages: (1) detecting attention sink tokens through hidden-state spike analysis, (2) selecting grounding heads that exhibit cross-modal imbalance, and (3) redistributing attention from sink tokens to instruction tokens.}
    \label{fig:arch}
\end{figure}

\section{IGAR: Instruction-Guided Attention Recalibration}
\label{sec:method}

As illustrated in Fig.~\ref{fig:arch}, we propose \textbf{I}nstruction-\textbf{G}uided \textbf{A}ttention \textbf{R}ecalibration (\textbf{IGAR}), a train-free inference-time mechanism that restores linguistic grounding in Vision-Language-Action (VLA) models by correcting vision-dominant attention sink imbalance during action prediction. We observe that, in modern VLA architectures, action-query tokens disproportionately attend to visually salient tokens, forming vision-dominant attention sinks that suppress instruction tokens even when the instruction semantically constrains the action. We attribute linguistic blindness to a structural cross-modal competition imbalance induced by attention sink dynamics. IGAR addresses this imbalance by detecting attention sinks through hidden-state spike analysis, selectively targeting cross-modal fusion heads, and recalibrating attention mass toward instruction tokens during inference. Our method operates entirely within the forward pass, requires no gradient updates or retraining, does not modify model parameters, and can be seamlessly integrated into transformer-based VLA backbones without architectural changes, making it suitable as a plug-and-play module for deployed robotic policies.

\subsection{Problem Formulation}
\label{sec:problem}

\noindent \textbf{Instruction Contradiction Setting.}
A VLA model $f_\theta$ maps observation $\bm{o}_t$ and a language instruction $\ell$ to action $\bm{a}_t = f_\theta(\bm{o}_t, \ell)$. Under ICBench, we evaluate the model using a semantically contradictory instruction $\tilde{\ell}$, while keeping the observation unchanged. By construction,  $\tilde{\ell}$ is incompatible with the visual scene and cannot be satisfied through valid task execution. This setup isolates the role of language instruction grounding in action generation. The diagnostic evaluation is particularly important for safety-critical robotic deployment, where blindly executing an impossible instruction may lead to unsafe behaviors. A policy that genuinely integrates instruction semantics should fail or abstain under contradiction. A policy that relies primarily on visual priors may still succeed despite the semantic inconsistency.

\noindent \textbf{Linguistic Grounding Score.} Let $\mathrm{SR}(f_\theta, {\ell})$ denote the task success rate of policy $f_\theta$ when executing under standard instruction ${\ell}$.

We define the Linguistic Grounding Score (LGS) as: 

\begin{equation}
    \mathrm{LGS}(\tilde{\ell}) = \mathrm{SR}(f_\theta, {\ell}) - \mathrm{SR}(f_\theta, \tilde{\ell}),
\end{equation}
where $\mathrm{SR}(f_\theta, \tilde{\ell})$ is the success rate respect to the standard instruction when executing the task using contradictory instruction  $\tilde{\ell}$. As a result, a perfect grounded model should fail under contradiction, yielding $\mathrm{LGS}(\tilde{\ell})=\mathrm{SR}(f_\theta, {\ell})$. A linguistically insensitive model that ignores instruction semantics will succeed regardless of contradiction, yielding $\mathrm{LGS}{\approx}0$.

\subsection{Instruction-Guided Attention Recalibration}
\label{sec:calm}

IGAR operates within the model’s forward pass by selectively recalibrating attention distributions in the transformer layers. Rather than globally modifying attention weights, IGAR intervenes only in structurally imbalanced heads identified through attention sink analysis. The procedure consists of three stages: (1) attention sink token detection, (2) grounding head selection, and (3) attention redistribution.

\noindent \textbf{Attention Sink Token Detection.} Let $\bm{H} \in \mathbb{R}^{N \times D}$ denote the hidden states from an intermediate transformer layer, where $N$ is the sequence length and $D$ the hidden dimension. 

We first compute the RMS norm per token:
\begin{equation}
    r_i = \sqrt{\frac{1}{D}\sum_{d=1}^{D} H_{i,d}^2}, \quad i = 1, \ldots, N.
\end{equation}
However, large token norms alone do not necessarily indicate structural dominance. To detect localized extreme activations, which is characteristic of attention sinks, we introduce a spike ratio per feature dimension:

\begin{equation}
    \phi(d) = \frac{\max_{i} |H_{i,d}|}{\mathrm{mean}_{i} |H_{i,d}| + \epsilon},
    \label{eq:spike}
\end{equation}
where $\epsilon$ is a small constant for numerical stability.
We select the top-$k$ dimensions ($k{=}5$) from the set $\mathcal{D}_{\text{spike}} = \{d : \phi(d) > \gamma\}$ with $\gamma{=}3.0$.
This spike criterion ensures that only dimensions with \textit{localized} extreme activations (few tokens with high values, not all tokens uniformly high) are selected, avoiding false-positive sink identification. A token $i$ is then classified as a \textbf{sink token} if:
\begin{equation}
    \mathcal{S} = \left\{i : \max_{d \in \mathcal{D}_{\text{spike}}} |H_{i,d}| > \tau \right\}, \quad \tau = 20.
    \label{eq:sink}
\end{equation}
We partition sinks into visual sinks $\mathcal{S}_\mathcal{V} = \mathcal{S} \cap \mathcal{V}$ and text sinks $\mathcal{S}_\mathcal{T} = \mathcal{S} \cap \mathcal{T}$.

\subsubsection{Grounding Head Selection.}
Not all attention heads are equally susceptible to sink distortion.
For each head $h$ in layer $l$ and each query position $q$ beyond the image region, we evaluate two conditions:
\begin{align}
    c_1(h,q) &: \frac{\sum_{j \in \mathcal{S}_\mathcal{V}} A^{h}_{q,j}}{\sum_{j \in \mathcal{V}} A^{h}_{q,j} + \epsilon} \leq \rho, \label{eq:cond1} \\
    c_2(h,q) &: \sum_{j \in \mathcal{V}} A^{h}_{q,j} \geq \alpha. \label{eq:cond2}
\end{align}
A head-query pair $(h,q)$ is selected for reallocation if both conditions hold, with $\rho{=}0.4$ and $\alpha{=}0.01$.
Condition~$c_1$ ensures the head is not already dominated by visual sinks (structural attention rather than semantic), while $c_2$ ensures the head allocates meaningful attention to visual tokens (filtering out text-only heads).
When no visual sinks are detected (\ie, $\mathcal{S}_\mathcal{V}{=}\emptyset$), $c_1$ is trivially satisfied and selection proceeds on $c_2$ alone.

\subsubsection{Attention Redistribution.}
For each selected head-query pair \((h,q)\), IGAR redistributes attention from sink tokens to non-sink tokens by first scaling down the attention of sink tokens by a factor \(p=0.6\). This operation frees a total budget
\begin{equation}
    \Omega^{h}_{q} = (1-p) \sum_{j \in \mathcal{S}_{\mathcal{T}}} A^{h}_{q,j}
\end{equation}
which is then reallocated to non-sink tokens proportionally to their original attention weights:
\begin{equation}
{A'}^{h}_{q,j} = A^{h}_{q,j} + \Omega^{h}_{q} \cdot \frac{A^{h}_{q,j}}{\sum_{j' \in \mathcal{T}_{\text{ns}}} A^{h}_{q,j'} + \epsilon}, \quad \forall\, j \in \mathcal{T}_{\text{ns}}
\end{equation}
where \(\mathcal{T}_{\text{ns}} = \mathcal{T} \setminus \mathcal{S}_{\mathcal{T}}\) denotes the set of non-sink text tokens.

\section{Experiments}
\label{sec:experiments}

We evaluate the proposed Instruction-Guided Attention Recalibration (IGAR) under the controlled OOD instruction contradiction setting introduced by ICBench to answer three key questions:
\begin{itemize}
    \item \textbf{Q1:} Do current Vision-Language-Action (VLA) models exhibit linguistic blindness under contradictory instructions?
    \item \textbf{Q2:} Can IGAR restore language grounding during action generation?
    \item \textbf{Q3:} Does IGAR preserve performance on valid in-distribution tasks?
\end{itemize}

\subsection{Experimental Setup}
\label{sec:exp_setup}

\textbf{VLAs and Benchmark.} We evaluate three representative VLA architectures ($\pi_0$~\cite{pi0}, $\pi_{0.5}$~\cite{pi05}, and OpenVLA-OFT~\cite{openvla_oft}) across 30 simulated robot manipulation tasks from the LIBERO benchmark~\cite{libero}, covering \textit{Spatial}, \textit{Object}, and \textit{Goal} suites. For each task variant, we perform 50 independent rollouts. These environments cover diverse robotic manipulation settings, including spatial reasoning, object manipulation, and goal-conditioned tasks. Our ICBench is constructed by introducing controlled contradictory instructions into the LIBERO tasks.

\noindent \textbf{Evaluation Metrics.} We report two primary evaluation metrics. Task Success Rate (SR) measures the percentage of successful task completions. Under standard instructions, a high SR indicates correct task execution and is therefore desirable. However, under the ICBench contradiction setting, a high SR becomes undesirable, as it implies that the policy completes the task despite contradictory instructions, suggesting that the model relies primarily on visual priors rather than instruction semantics. To quantify the influence of language instructions on action generation, we adopt the Linguistic Grounding Score (LGS) defined in Sec.~\ref{sec:problem}. LGS measures the performance gap between executions under valid instructions and contradictory instructions, with higher values indicating stronger reliance on linguistic instructions.

\noindent \textbf{Implementation Details.}
\method{} is applied uniformly at inference time without any task-specific tuning. For attention-based decoders ($\pi_{0.5}$ and OpenVLA-OFT), interventions span the initial 16 layers using unified constraints for visual-sink portion boundaries ($\rho{=}0.4$) and text-sink decay ($p{=}0.6$), alongside fixed detection thresholds ($\tau{=}20$, $\gamma{=}3.0$). For $\pi_0$'s continuous flow-matching diffusion, we deploy a discrete contrastive formulation that restricts divergence triggers ($\tau_{\mathrm{detect}}{=}0.15$) and interpolation boundaries ($\tau_{\mathrm{max}}{=}0.50$). This standardized configuration isolates the inherent robustness of \method{} across fundamentally disparate generation paradigms.

\subsection{Diagnosing Linguistic Blindness}
\label{sec:main_results}

Table~\ref{tab:vulnerability} presents the VLA baseline evaluation under ICBench, comparing performance under normal instructions and contradictory instructions. Across all three VLAs, we observe a consistent pattern: modern VLA models frequently ignore contradictory language inputs and continue executing visually plausible actions, revealing a systemic form of \textit{linguistic blindness}. First, models often maintain extremely high success rates even when the instruction is semantically invalid. Across the \textit{Spatial} and \textit{Object} suites, success rates frequently exceed 90\% under contradictory instructions, particularly for $\pi_{0.5}$ and OpenVLA-OFT. This indicates that the policy continues to complete the task by relying on visual cues rather than instruction semantics. As a consequence, the resulting LGS remain very low, suggesting that language contributes minimally to action generation in these scenarios. Second, the severity of linguistic blindness varies across architectures. $\pi_0$ shows relatively higher LGS values compared to $\pi_{0.5}$ and OpenVLA-OFT, indicating slightly stronger language grounding. In contrast, $\pi_{0.5}$ often exhibits low LGS values, implying that contradictory instructions have almost no influence on the executed actions. These results demonstrate that despite their multimodal design, current VLA models often rely on vision-dominant execution strategies. Language instructions exert limited influence, though they directly constrain the physical feasibility of the action, highlighting a critical vulnerability for deploying VLA systems in real-world settings.

\begin{table}[h]
\centering
\caption{\textbf{Linguistic Blindness Diagnosis on ICBench.}
Success Rate (SR, \%) and Linguistic Grounding Score (LGS) under normal and contradictory instructions. High SR and low LGS under contradiction (V1-V4) indicate that the policy ignores instruction semantics and follows visual priors.
\tmark{Highlighted cells} denote SR $\geq$\,90\%, revealing severe linguistic blindness.}
\label{tab:vulnerability}
\small
\setlength{\tabcolsep}{3.3pt}
\begin{tabular}{ll|cc|cc|cc}
\toprule
& & \multicolumn{2}{c|}{$\bm{\pi_0}$} & \multicolumn{2}{c|}{$\bm{\pi_{0.5}}$} & \multicolumn{2}{c}{\textbf{OpenVLA-OFT}} \\
\textbf{Suite} & \textbf{Instruction} & SR & LGS & SR & LGS & SR & LGS \\
\midrule
\multirow{5}{*}{\textit{Spatial}}
 & Normal & 96.8 & -- & 97.4 & -- & 97.6 & -- \\
 & V1 (operand attribute) & \tmark{90.4} & 6.4 & \tmark{96.2} & 1.2 & \tmark{97.8} & -0.2 \\
 & V2 (target attribute) & \tmark{96.2} & 0.6 & \tmark{97.8} & -0.4 & \tmark{96.4} & 1.2 \\
 & V3 (dual attribute) & 89.8 & 7.0 & \tmark{96.4} & 1.0 & \tmark{96.2} & 1.4 \\
 & V4 (spatial relation) & \tmark{92.4} & 4.4 & \tmark{97.6} & -0.2 & \tmark{92.4} & 5.2 \\
\midrule
\multirow{5}{*}{\textit{Object}}
 & Normal & 98.8 & -- & 98.4 & -- & 98.4 & -- \\
 & V1  (operand attribute) & \tmark{93.8} & 5.0 & \tmark{96.2} & 2.2 & \tmark{97.8} & 0.6 \\
 & V2 (target attribute) & \tmark{90.2} & 8.6 & \tmark{90.4} & 8.0 & \tmark{96.2} & 2.2 \\
 & V3 (dual attribute) & \tmark{98.2} & 0.6 & \tmark{96.4} & 2.0 & \tmark{99.8} & -1.4 \\
 & V4 (spatial relation) & \tmark{91.6} & 7.2 & 88.2 & 10.2 & \tmark{99.6} & -1.2 \\
\midrule
\multirow{5}{*}{\textit{Goal}}
 & Normal & 95.8 & -- & 97.6 & -- & 98.0 & -- \\
 & V1  (operand attribute) & \tmark{90.2} & 5.6 & \tmark{93.8} & 3.8 & \tmark{97.8} & 0.2 \\
 & V2 (target attribute) & 84.4 & 11.4 & \tmark{90.4} & 7.2 & \tmark{98.2} & -0.2 \\
 & V3 (dual attribute) & 88.2 & 7.6 & \tmark{94.2} & 3.4 & \tmark{98.4} & -0.4 \\
 & V4  (spatial relation) & 76.4 & 19.4 & \tmark{93.6} & 4.0 & \tmark{90.2} & 7.8 \\
\bottomrule
\end{tabular}
\end{table}

\subsection{IGAR Restores Linguistic Grounding}
\label{sec:gaseraser_results}
\begin{figure}[h]
    \centering
    \includegraphics[width=0.9\textwidth]{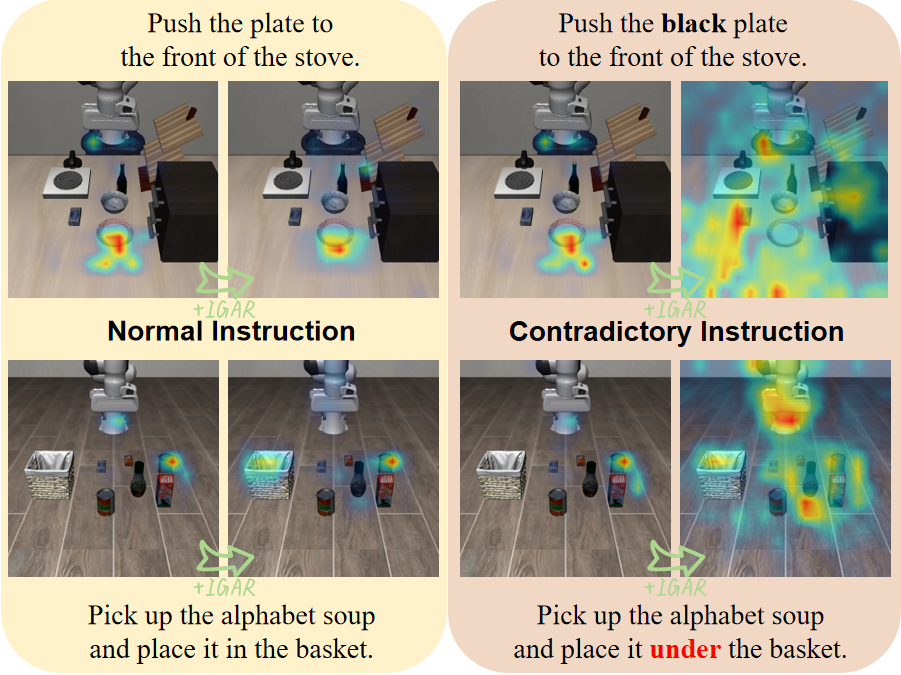}
    \caption{\textbf{Attention visualization of OpenVLA-OFT with and without IGAR.} We visualize cross-modal attention maps under normal and contradictory instructions. The baseline policy attends primarily to salient regions regardless of instruction semantics, while IGAR redistributes attention toward instruction-relevant objects and spatial regions, mitigating visual attention sinks and improving linguistic grounding.}
    \label{fig:heatmap}
\end{figure}
Table~\ref{tab:mitigation} reports the performance of VLA models under contradictory instructions after applying IGAR. Compared with the baseline results in Table~\ref{tab:vulnerability}, IGAR consistently reduces erroneous task execution while substantially increasing the LGS, indicating that action generation becomes more sensitive to instruction semantics.
First, IGAR effectively suppresses visually driven execution under contradictory instructions. Across all suites, SR drops significantly compared to the baseline setting, particularly for the $\pi_0$ and OpenVLA-OFT. For example, in the \textit{Goal} suite, SR decreases to as low as 36.4\% under spatial contradictions (V4), while LGS increases dramatically to 59.4. This indicates that the policy refrains from executing actions that violate instruction semantics, demonstrating restored language grounding. Second, the improvement varies across model architectures and suites. The largest gains are observed for $\pi_0$, where LGS consistently exceeds 40 and reaches up to 59.4 in goal suites. OpenVLA-OFT also shows substantial improvements, achieving LGS values above 30 in several goal tasks. In contrast, the $\pi_{0.5}$ model exhibits more limited gains, showing strong reliability on visual cues. Overall, the results demonstrate that IGAR effectively restores the influence of language instructions during action generation. By redistributing attention toward instruction tokens, the method mitigates vision-dominant biases and enables VLA policies to better detect and respond to semantically inconsistent instructions.

\begin{table}[h]
\centering
\caption{\textbf{\method{} mitigation under ICBench.} We evaluate VLA models under contradictory instructions after applying IGAR. SR measures task completion despite contradiction (lower is better), while LGS measures the influence of language instructions (higher is better). \gmark{Green cells} indicates LGS $\geq$\,10.}
\label{tab:mitigation}
\small
\setlength{\tabcolsep}{3.2pt}
\begin{tabular}{ll|cc|cc|cc}
\toprule
& & \multicolumn{2}{c|}{$\bm{\pi_0}$} & \multicolumn{2}{c|}{$\bm{\pi_{0.5}}$} & \multicolumn{2}{c}{\textbf{OpenVLA-OFT}} \\
\textbf{Suite.} & \textbf{Contradiction} & SR & LGS & SR & LGS & SR & LGS \\
\midrule
\multirow{4}{*}{\rotatebox{90}{\textit{Spatial}}}
 & V1 & 76.4 & \gmark{20.4} & 95.8 & 1.6 & 86.4 & \gmark{11.2} \\
 & V2 & 84.2 & \gmark{12.6} & 93.6 & 3.8 & 86.2 & \gmark{11.4} \\
 & V3 & 80.4 & \gmark{16.4} & 94.8 & 2.6 & 90.2 & 7.4 \\
 & V4 & 76.2 & \gmark{20.6} & 99.6 & -2.2 & 88.4 & 9.2 \\
\midrule
\multirow{4}{*}{\rotatebox{90}{\textit{Object}}}
 & V1 & 88.2 & \gmark{10.6} & 94.2 & 4.2 & 88.4 & \gmark{10.0} \\
 & V2 & 86.4 & \gmark{12.4} & 90.4 & 8.0 & 84.2 & \gmark{14.2} \\
 & V3 & 93.6 & 5.2 & 87.6 & \gmark{10.8} & 88.2 & \gmark{10.2} \\
 & V4 & 90.4 & 8.4 & 88.4 & \gmark{10.0} & 82.4 & \gmark{16.0} \\
\midrule
\multirow{4}{*}{\rotatebox{90}{\textit{Goal}}}
 & V1 & 46.4 & \gmark{49.4} & 90.2 & 7.4 & 66.4 & \gmark{31.6} \\
 & V2 & 40.2 & \gmark{55.6} & 82.8 & \gmark{14.8} & 70.2 & \gmark{27.8} \\
 & V3 & 46.2 & \gmark{49.6} & 92.4 & 5.2 & 64.2 & \gmark{33.8} \\
 & V4 & 36.4 & \gmark{59.4} & 96.2 & 1.4 & 58.4 & \gmark{39.6} \\
\bottomrule
\end{tabular}
\end{table}

\subsection{Baseline Preservation Under \method{}}
\label{sec:preservation}
Table~\ref{tab:preservation} evaluates whether IGAR affects performance under normal (non-contradictory) instructions. We compare the VLA baseline success rates with results obtained when IGAR is applied to the same tasks. Overall, IGAR preserves baseline performance with only marginal impact. Across all suites, the $\pi_0$ model shows negligible performance changes, with an average decrease of only 0.4\%. Similarly, OpenVLA-OFT maintains almost identical performance, with an average change of +0.5\%. These results indicate that the attention recalibration introduced by IGAR does not interfere with correct instruction following when the language input is consistent with the visual scene. 
Furthermore, we measure per-step inference latency on a single NVIDIA RTX 5090 GPU over 50 trials. For OpenVLA-OFT, latency increases from $98.7 \pm 2.0$ ms to $105.6 \pm 0.5$ ms, corresponding to a modest $+7.0$ ms overhead.

\begin{table}[h]
\centering
\caption{\textbf{Baseline preservation under \method{}.} Success Rate (SR) comparison under normal (non-contradictory) instructions for baseline and baseline with our \method{}. "OFT" refers to "OpenVLA-OFT".}

\label{tab:preservation}
\small
\setlength{\tabcolsep}{3.8pt}
\begin{tabular}{l|cc|cc|cc}
\toprule

\textbf{Suites} & $\bm{\pi_0}$ & $\bm{\pi_0}$+\method{} & $\bm{\pi_{0.5}}$ & $\bm{\pi_{0.5}}$+\method{} & \textbf{OFT} & \textbf{OFT}+\method{} \\
\midrule
\textit{Spatial} & 96.8 & 96.4\,{\scriptsize($-$0.4)} & 97.4 & \gmark{98.2}\,{\scriptsize(+0.8)} & 97.6 & 97.6\,{\scriptsize(+0.0)} \\
\textit{Object} & 98.8 & 98.2\,{\scriptsize($-$0.6)} & 98.4 & 94.4\,{\scriptsize($-$4.0)} & 98.4 & \gmark{99.8}\,{\scriptsize(+1.4)} \\
\textit{Goal} & 95.8 & 95.6\,{\scriptsize($-$0.2)} & 97.6 & 96.4\,{\scriptsize($-$1.2)} & 98.0 & \gmark{98.2}\,{\scriptsize(+0.2)} \\
\midrule
\textit{Average} & 97.1 & 96.7\,{\scriptsize($-$0.4)} & 97.8 & 96.3\,{\scriptsize($-$1.5)} & 98.0 & 98.5\,{\scriptsize(+0.5)} \\
\bottomrule
\end{tabular}
\end{table}

\subsection{Ablation Study and Hyperparameter Sensitivity}
\label{sec:ablation}

\begin{figure}[h]
    \centering
    \includegraphics[width=0.9\textwidth]{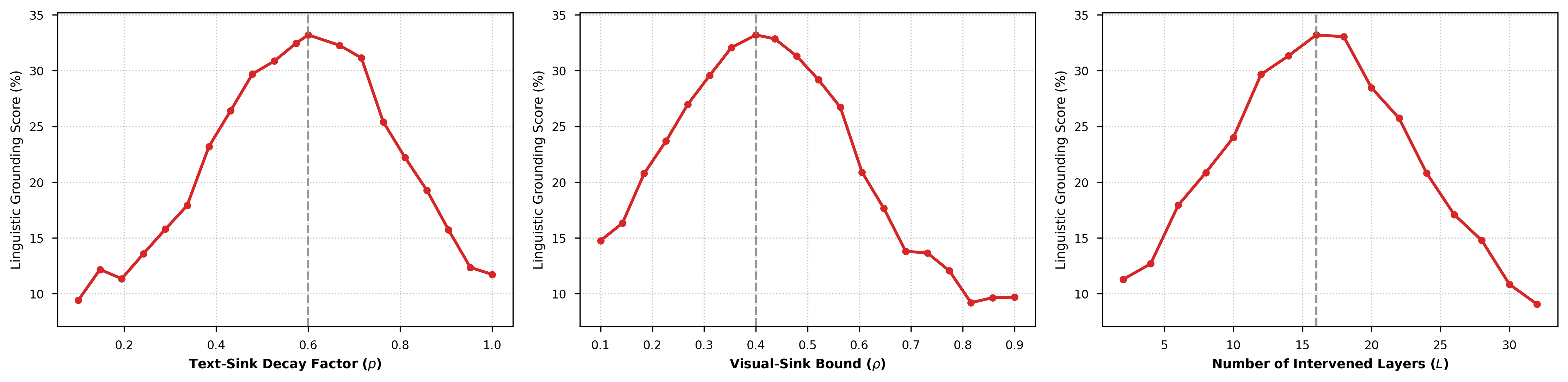}
 \caption{\textbf{Hyperparameter Sensitivity.} Impact of the text-sink decay factor ($p$), head selection bound ($\rho$), and number of intervened layers ($L$) in terms of linguistic grounding performance (LGS). Results are reported using the OpenVLA-OFT architecture on the libero\_goal benchmark suite. The dashed lines indicate the selected values.}
    \label{fig:hyperparam}
\end{figure}

We systematically analyze the influence of key hyperparameters governing \method{}'s intervention constraints using OpenVLA-OFT on the \texttt{libero\_goal} benchmark. As detailed in Figure~\ref{fig:hyperparam}, we evaluate the LGS to measure the effectiveness of language-action grounding. Applying an overly aggressive text-sink decay factor ($p<0.3$) or an insufficient one ($p>0.8$) disrupts the optimal reinforcement of instructional tokens, directly reducing the overall LGS. Similarly, excessively loose visual-sink bounds ($\rho>0.6$) fail to adequately intercept structural visual heads responsible for attention sinks, while overly strict boundaries interfere with normal processing. Furthermore, intervening across too many transformer layers ($L>24$) tends to corrupt deep representations that are highly specialized for continuous action regression, whereas intervening on too few layers fails to provide a strong enough corrective signal. Our default configurations ($p=0.6$, $\rho=0.4$, $L=16$) strategically target the mid-level layers where cross-modal semantic fusion primarily occurs, fully maximizing the LGS and effectively mitigating linguistic blindness.

We further ablate the three stages of IGAR using $\pi_{0.5}$
on LIBERO-Goal under ICBench V1. As shown in Table~\ref{tab:component_ablation}, full IGAR increases attention-to-text from 0.08 to 0.15 and improves LGS from 3.8 to 7.4. Removing sink detection or head selection weakens the effect, while detection alone leaves both attention-to-text and LGS unchanged. These results verify that sink detection, grounding-head selection, and attention redistribution are complementary rather than redundant.

\begin{table}[h]
\centering
\caption{\textbf{Component ablation of IGAR.} Results are reported using $\pi_{0.5}$ on LIBERO-Goal under ICBench V1.}
\label{tab:component_ablation}
\small
\resizebox{0.8\linewidth}{!}{
\begin{tabular}{lccccc}
\toprule
Config & Sink Det. & Head Sel. & Redist. & Attn-to-Text ($\uparrow$) & LGS ($\uparrow$) \\
\midrule
Baseline ($\pi_{0.5}$) & -- & -- & -- & 0.08 & 3.8 \\
Full IGAR & $\checkmark$ & $\checkmark$ & $\checkmark$ & \textbf{0.15} & \textbf{7.4} \\
\quad w/o Sink Detection & -- & $\checkmark$ & $\checkmark$ & 0.11 & 4.6 \\
\quad w/o Head Selection & $\checkmark$ & -- & $\checkmark$ & 0.10 & 4.2 \\
\quad Detection only & $\checkmark$ & -- & -- & 0.08 & 3.8 \\
\bottomrule
\end{tabular}
}
\end{table}

\subsection{Real-World Evaluation}
\begin{figure}[h]
    \centering
    \includegraphics[width=0.9\textwidth]{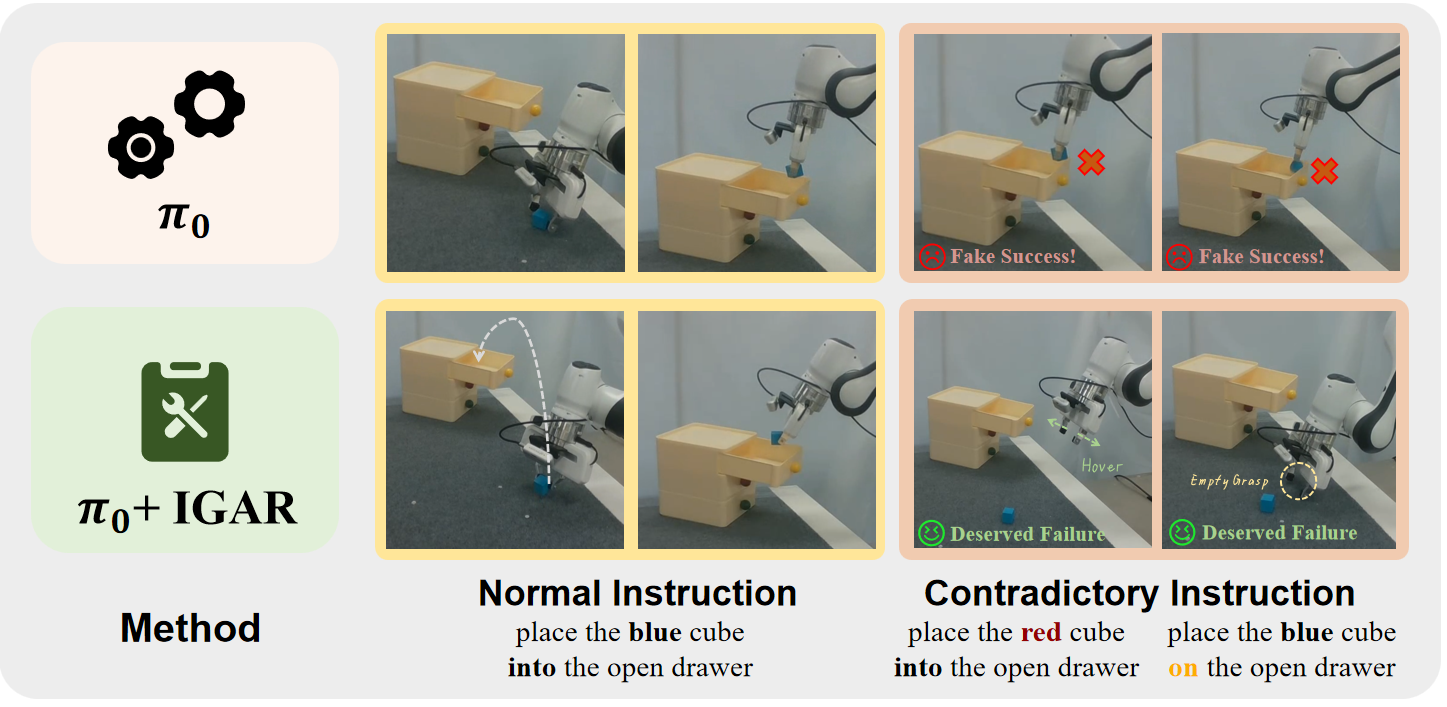}
    \caption{\textbf{Real-world experiments.} We test the $\pi_0$ policy with and without IGAR on the task "placing the blue cube into the open drawer". Under normal instructions (left), both policies successfully place the blue cube into the open drawer. Under contradictory instructions (right), the $\pi_0$ policy still executes a visually plausible trajectory and produces a \emph{fake success}. In contrast, IGAR restores linguistic grounding and prevents incorrect task execution, resulting in a \emph{deserved failure}.}
    \label{fig:realworld}
\end{figure}

We further validate IGAR on a real robotic platform to examine whether the linguistic grounding improvements observed in simulation transfer to physical manipulation. As shown in Fig.~\ref{fig:realworld}, experiments are conducted on a Franka Research 3 robotic arm equipped with two Intel RealSense D435 cameras, one mounted on the wrist and another providing a third-person viewpoint. We evaluate the $\pi_0$ policy on a cube placement task: \emph{``place the blue cube into the open drawer''}. Under normal instructions, both the $\pi_0$ policy and $\pi_0$+IGAR reliably complete the task, confirming that IGAR preserves the original policy behavior when the instruction is consistent with the scene. However, when the instruction becomes contradictory (e.g., requesting a non-existent object attribute or an impossible spatial relation), the baseline policy frequently continues executing the visually plausible trajectory and completes the manipulation despite the semantic inconsistency. This behavior results in \emph{fake success}, where the robot performs the correct physical action but violates the instruction semantics. In contrast, IGAR restores the influence of language during action generation. Given contradictory instructions, the recalibrated policy refrains from completing the task and instead produces safe behaviors such as hovering or empty grasp attempts, effectively interrupting the manipulation. These outcomes represent \emph{deserved failures}, indicating that the policy correctly detects the instruction inconsistency.

\section{Conclusion}
\label{sec:conclusion}

In this paper, we have revealed a critical reliability issue in Vision-Language-Action (VLA) models that we term \textit{linguistic blindness}, where policies prioritize visual priors over instruction semantics during action generation. To systematically diagnose this failure mode, we introduce ICBench, a controlled benchmark that evaluates language–action grounding under structured contradictory instructions. We further propose Instruction-Guided Attention Recalibration (IGAR), a train-free inference-time intervention that restores linguistic grounding by redistributing attention from sink tokens to instruction tokens. Experiments on three representative VLA architectures demonstrate that modern models often execute visually plausible actions even when instructions are logically inconsistent with the scene, while IGAR effectively mitigates such failures without degrading baseline task performance.

\section*{Acknowledgment}
This project is supported by the Ministry of Education (MOE), Singapore, under its Academic Research Fund (AcRF) Tier 2 (Proposal ID: T2EP20125-0048). Any opinions, findings and conclusions or recommendations expressed in this material are those of the authors and do not reflect the views of the Ministry of Education, Singapore. This project was also supported by NSFC Project (No. 62522206) and Tsinghua University Initiative Scientific Research Program (Student Academic Research Advancement Program).

\bibliographystyle{splncs04}
\bibliography{main}
\end{document}